# NEXT LEVEL OF DATA FUSION FOR HUMAN FACE RECOGNITION

M. K. Bhowmik[a], G. Majumder[a], D. Bhattacharjee[b], D. K. Basu[b], M. Nasipuri[b]

[a]Department of Computer Science & Engineering, Tripura University (A Central University), Suryamaninagar.
[b]Department of Computer Science & Engineering, Jadavpur University, Kolkata.

## ABSTRACT

This paper demonstrates two different fusion techniques at two different levels of a human face recognition process. The first one is called data fusion at lower level and the second one is the decision fusion towards the end of the recognition process. At first a data fusion is applied on visual and corresponding thermal images to generate fused image. Data fusion is implemented in the wavelet domain after decomposing the images through Daubechies wavelet coefficients (db2). During the data fusion maximum of approximate and other three details coefficients are merged together. After that Principle Component Analysis (PCA) is applied over the fused coefficients and finally two different artificial neural networks namely Multilayer Perceptron(MLP) and Radial Basis Function(RBF) networks have been used separately to classify the images. After that, for decision fusion based decisions from both the classifiers are combined together using Bayesian formulation. For experiments, IRIS thermal/visible Face Database has been used. Experimental results show that the performance of multiple classifier system along with decision fusion works well over the single classifier system.

## INTRODUCTION

Biometric security systems based on human face recognition have already been an established field for authentication and surveillance purposes. Till date, no feature extraction technique for human face recognition exists which is capable to ease the work of classifiers to large extent because if features are extracted with highest accuracy then the task of classifier becomes very easy, a simple distance measure e.g. Euclidean distance may suffice. But that doesn't happen in reality and we need to design complex classifiers. Having such complex classifier it has been observed that in most of the cases to classify complex pattern like human faces we take the advantage of multiple classifiers by combining their individual result into a final classification answer. In order to achieve such an objective the ultimate goal of this work is to design a multi-classifier system for human face recognition. Several works have already been developed using fusion of decisions from different classifier; like for land cover [15-17], sea ice classification [18] cloud classification [19], and also for face recognition. There are mainly three types of fusion strategies [12], namely, information/data fusion (low-level fusion), feature fusion (intermediate-level fusion), and decision fusion (high-level fusion). Data fusion combines several sources of raw data to produce new raw data that is expected to be more informative and synthetic than the inputs [12]. Decision fusion uses a set of classifiers to provide a better and unbiased result. The classifiers can be of same or different type and can also have same or different feature sets [13]. Hence a set of classifiers is used and finally the outputs of all the classifiers are merged together by various methods to obtain the final output. In recent years, decision fusion techniques have been investigated [9] [13] [10] and their application on classification domain has been widely tested. Decision fusion can be defined as the process of fusion information of individual data sources. The main contribution of this paper is a decision fusion of two different artificial neural networks based on a distance the individual test image and corresponding feature images. First the individual images trained separately using two different classifier and finally decision of these two classifiers have combined together. There are many techniques have been developed for face recognition given in Table 1.

J. Czyz et al. [20], used decision fusion to demonstrate their fully automatic multi-frame–multi-expert system on realistic database to. They study sequential fusion, that is, the fusion of outputs of a single face authentication algorithm obtained on several video frames. This type of fusion is referred to as intra-modal fusion. The main contribution of their paper is a fusion architecture which takes into account the distinctive features of the intra-modal and sequential fusion. They used two different face verification algorithms, namely a Linear

**Table 1: Existing Techniques of decision fusion for Human face recognition**

| Authors | Technique | Reference |
|---|---|---|
| J. Czyz et al. | Decision Fusion for Face Authentication | [20] |
| J. Heo et al. | Robust Face Recognition | [21] |
| B. Gokberk et al. | Rank-based Decision Fusion | [22] |
| B. Gokberk et al. | Decision Level Fusion for 3D Face Recognition | [23] |

Discriminant Analysis (LDA) based and a Support Vector Machine (SVM) based algorithm. J. Heo et al. [21], use data and decision fusion for robust face recognition of visual and thermal images. Data Fusion produces an illumination-invariant face image by detecting the eyeglass, which blocks the thermal energy, and replaced them with an eye template from visual and thermal face images, and Decision Fusion combines matching scores of individual face recognition modules. They determine the matching score of Decision Fusion as a weighted sum of visual recognition module and thermal recognition module, or the largest one of them. They used the NIST/Equinox database used for evaluation of fusion-based face recognition performances. Their comparison result show that fusion-based face recognition techniques outperformed individual visual and thermal face recognizers under illumination variation and facial expressions. B. Gokberk et al. [22], divided their work into two parts: In the first part, they have developed face classifiers which use the techniques based on point clouds, surface normal's, facial profiles, and statistical analysis of depth images. They used 3D RMA dataset for their study, and the Experimental results show that the Linear Discriminant Analysis (LDA) based representation of depth images and point cloud representation perform best. In the second part, two different multiple-classifier architectures are developed to fuse individual shape-based face recognizers in parallel and hierarchical fashions at the decision level. It is shown that a significant performance improvement is possible when using rank-based decision fusion in ensemble methods. In another case [23], they formulate several 3D face recognition algorithms and evaluate their comparative performance. They address these three basic issues of face data representation are, feature selection and classifier fusion. In the second tier, they investigate a fairly complete list of features, to be extracted from each face representation. These features are either of local geometric variety, like surface curvature, or of global variety, like transform techniques. Several individual classifiers, called also face experts, have resulted from the judicious combination of face representations and of face features.

Rest of the paper has been organized as follows: the overview of the system is discussed next. After that experimental results and discussions are given and finally conclusions relating to this work are given.

## SYSTEM OVERVIEW

The complete system implemented in this work is described with the help of a block diagram, which is shown in Fig 1. In the first step, Discrete Wavelet Transform (DWT) is used for decomposition of both the thermal and visual images up to level five using wavelet namely Daubechies (db2). In the second step, synthesized fused image is generated by the combining of decomposed thermal and visual image by following data fusion scheme. In the third step, a feature selection algorithm is applied to calculate the features of the fused image to represent the images to lower dimension space. In fourth step two different classifiers are applied on the same image dataset and finally outcomes of these two classifiers have been combined together for implementing decision fusion which is a higher level fusion as compared to data fusion.

## Visual Image

Visual images are optical images, pertaining to a visual perception or can be referred to as a percept that arises from the eyes or as an image in the visual system [1]. The task of face recognition based only on the visible spectrum is still a challenging problem under uncontrolled environments. The challenges are even more profound when one considers the large variations in the visual stimulus due to illumination conditions and pose [2]. Such problems are quite unavoidable in applications such as outdoor access control and surveillance. Performance of visual face recognition is sensitive to variations in illumination conditions and usually degrades significantly when the lighting is dim or when it is not uniformly illuminating the face. Recently, researchers have investigated the use of thermal infrared face images for person identification to tackle illumination variation.

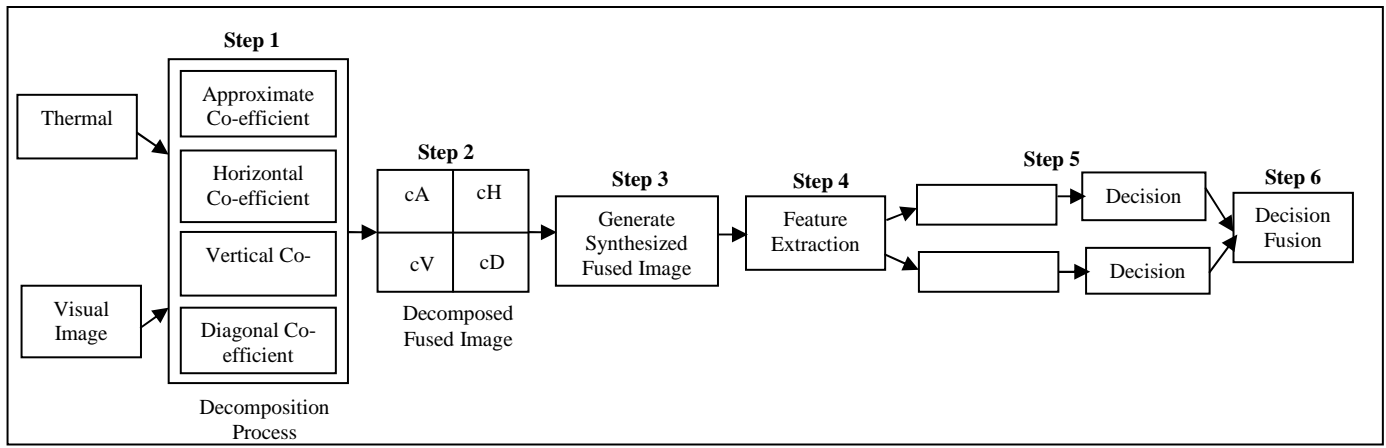

**Fig. 1: System block diagram**

## Thermal Image

Thermal IR imagery [3] has been suggested as a viable alternative in detecting disguised faces and handling situations where there is no control over illumination. Thermal IR images represent the heat patterns emitted from an object. Objects emit different amounts of IR energy according to their body temperature and characteristics. There are many advantages of thermal image like Face (and skin) detection, location, and segmentation are easier when using thermal images. It can work even in total darkness and also useful for detecting disguises. Over the different advantages it has certain disadvantages also like at the time of breathing it results big change in skin temperature and thermal images has low resolution.

## Fused Image

Task of interpreting images, either visual images alone or thermal images alone, is an unconstraint problem. The use of thermal data gathered by an infrared camera, along with the visual image, is seen as a way of resolving some of the difficulties present in case of visual and thermal faces. The technique of fusion combines the several sources of raw data (i.e. thermal and visual images in this case) to produce a new raw data which is expected to be more informative than the individuals.

## Data Fusion in wavelet domain

The main idea behind the fusion algorithm is i) the two images are to be processed and resample to the one with the same size; ii) they are respectively decomposed into the sub-images using forward wavelet transform, which have the same resolution at the same levels and different resolution among different levels (stage-I); iii) data fusion is performed based on the high-frequency sub-images of decomposed images (stage-II); and finally the result image is obtained using inverse wavelet transform (stage-III) [4-7], [12]. The process of data fusion is representing with the help of a block diagram in Fig. 2.

## Feature Selection

The task of the feature selection methods is to obtain the most relevant information from the original data and represent that information in a lower dimensionality space. Here, Principle Component Analysis (PCA) is used to calculate the feature of the 2D signals (i.e. face images) on the fused image of thermal and visual images. PCA tries to attain an optimal representation that minimizes the reconstruction error in a least-squares sense. Eigenvectors of the covariance matrix of the face images constitute the eigenfaces. The dimensionality of the face feature space is reduced by selecting only the eigenvectors possessing significantly large eigenvalues. Once the new face space is constructed, when a test image arrives, it is projected onto this face space to yield the feature vector—the representation coefficients in the constructed face space. The classifier decides for the identity of the individual, according to a similarity score between the test image's feature vector and the PCA feature vectors of the individuals in the database [8].

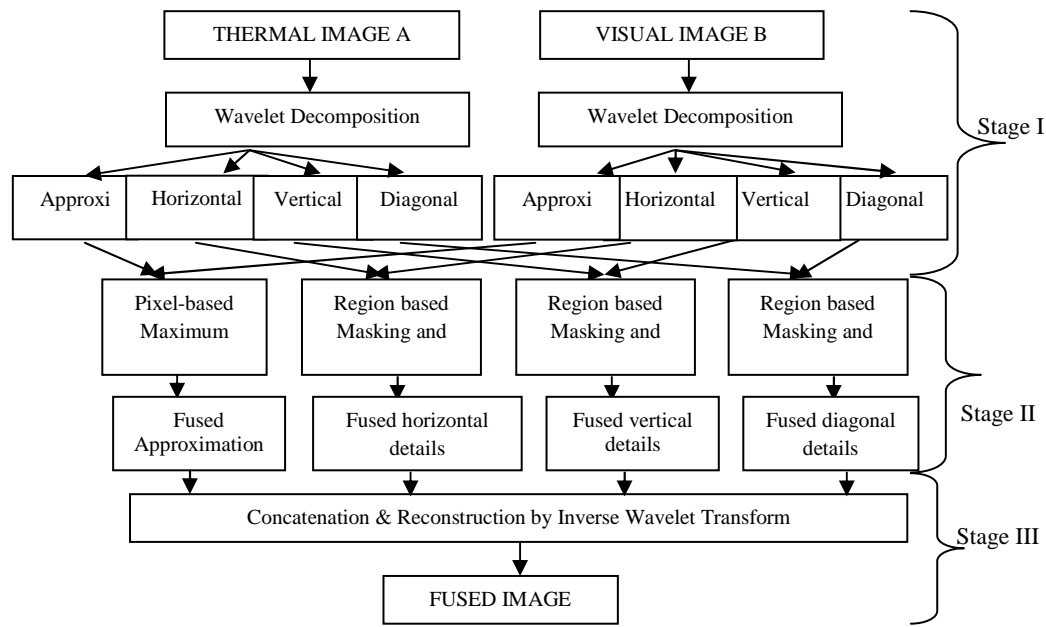

**Fig. 2: Steps of data fusion of images**

## Decision Fusion of Multiple Classifiers

Decision Fusion is a High-Level fusion technique, uses a set of classifiers on each output of statistical method (PCA in this case) to produce a better result [13]. The classifiers can be same or different type and can also have same or different feature sets. In face Recognition (image processing) classifier is a neural network used to training and testing the image data. There are different classifiers such as SVM with various kernels, ANN, KNN, GMM etc. Here two different artificial neural networks multilayer perceptron (MLP), and radial basis function (RBF) have been used to achieve decision fusion strategy for human face recognition. Decision Fusion uses a set of classifiers to improve the accuracy and efficiency of the classification system [11]. The three most important reasons to use multi-classifier is,

(a) To perform the efficiency of a face recognition system against large amount of data using a single classifier is not practical. So a multi classifier system will be an efficient approach, where data is partitioned into smaller subsets, trained with different classifier for different subsets and the output combined.
(b) A single classifier cannot perform well when the dataset of images varying between different classes. So a multiple classifiers with a subset of features may provide a better performance over a single classifier.
(c) Generalization performance also improved when a multiple classifier system is used.

Three different procedures are followed to achieve multiple classifiers system:

(i) **Variation of initial parameter of the classifiers:** A set of classifiers is created by varying the initial parameters but same training dataset.

(ii) **Variation of training dataset of the classifiers:** Different datasets used during training for individual classifiers.
(iii) **Variation in the number of individual classifiers used:** Two different classifiers MLP and RBF are used to classify the same image dataset.

A confusion matrix for a classifier is a matrix in which the actual class or a datum under test is represented by the matrix row, and the classification of that particular datum is represented by the confusion matrix column. The diagonal elements indicate correct classifications and, if the matrix is not normalized, the sum of row i is the total number of elements of class i that actually appeared in the data set. The element M[i][j] gives the number of times that a class i object was assigned to class j. For N classes the size of the confusion matrix is N × N+1, where the additional column for j = N+1 indicates the number of objects that are rejected. The conditional probability that a pattern y belong to class i (i.e. $C_i$) and classifier q assigns it to class j, denoted by, $F_q(y)=j$, can be given as [9]

$$P(y \in C_i | F_q(y) = j) = \frac{M^q[i][j]}{\sum_{i=1}^{N} M^q[i][j]} \quad (1)$$

Where, q = 1,2,.. Q and Q is the total number of classifiers. Here, Q = 2 for two classifiers viz. MLP and RBF; $M^q[i][j]$ is extension of confusion matrix for $q^{th}$ classifier.
Since, for Q classifiers classification results are $F_q(y)=j_k$ then a measure called belief value for the belongingness of y to class i can be given as[9]

$$bel(i) = \frac{\prod_{k=1}^{Q} P(y \in C_i | F_k(y) = j_k)}{\sum_{i=1}^{N} \prod_{k=1}^{Q} P(y \in C_i | F_k(y) = j_k)} \quad (2)$$

for i = 1, …,M.

For ant input face image y, it can be assigned to class i if bel(i) ≥ bel(j), ∀j ≠ I and bel(i) > γ. Otherwise, y is rejected to be a member of class i. It is also rejected when $F_q(y)=N+1$, for all q i.e. y has been rejected by all the classifiers. Here, γ is a parameter to control the performance of the decision fusion scheme discussed here. If the system to be deployed in critical environment, then in order to achieve higher degree of accuracy γ should be increased but obviously at the cost of recognition rate because if error rate is reduced then recognition rate must go down. An interesting result for this trade off can be found in [9]. In this experiment, the typical value for control parameter has been considered as γ = 0.95.

**EXPERIMENTAL RESULTS AND DISCUSSION**

Three separate experiments have been conducted in this work to demonstrate the advantages of using multiple classifiers over single classifier in face recognition system. Here, at first RBF and MLP as single classifiers have been investigated and then decisions from both the classifiers have been combined for decision fusion using equation(2) and the results of such investigations are given subsequently. For conducting experiments, in this work the IRIS (Imaging, Robotics and Intelligent System) face database has been considered. IRIS database simultaneously acquired unregistered thermal and visible face images under variable illuminations, expressions, and poses and its only one publicly available thermal and corresponding visual image dataset [14]. This work has been simulated using MATLAB 7 in a machine of the configuration 2.13GHz Intel Xeon Quad Core Processor and 16384.00MB of Physical Memory. All the visual, thermal, and fused images used in this experiment are of size 40×50 pixels.

**Training and Testing**

All three experiments has been conducted on same image set to showing how a multi-classifier system work better over a single classifier system for human face recognition. Total 220 visual and 220 thermal faces have been picked up from 10 different classes of the database (i.e. 22 thermal and 22 visual images per class) having the condition of severe illumination change. After that data fusion algorithm is applied over the images and generate 220 fused images. Among 220 fused images 110 images of 10 different classes (i.e. 11 images per class) have been used for training purpose and rest of the images are kept for testing purpose.

### Experimental Results on Multilayer Perceptron Neural Network

The dashed line of Fig. 3 represents the recognition rates using multilayer perceptron as a classifier. Among the 10 different classes no classes has achieved the maximum recognition rate. The highest recognition rate is 95.94% and the average recognition rate for the total 110 images is 78.72%.

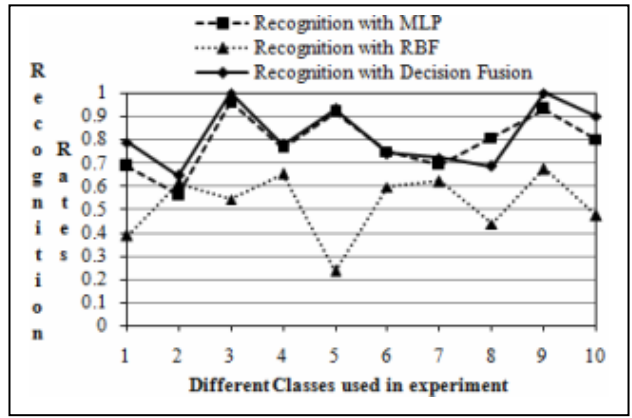

Fig 3: Recognition rates

### Experimental Results on Radial Basis Function Neural Network

The dotted line of figure 3 is the results of when radial basis function is used for classification. But for radial basis function degrades as compare to multilayer perceptron when number of test data is gradually increased. The highest recognition rate is 65.06% which is very less as compare to multilayer perceptron and the average recognition rate is 52.30% which is less than 25% from MLP.

### Experiment Results on Decision Fusion (Multiple Classifiers)

Finally a combined classifier is used for decision fusion and the results of decision fusion is represent using a solid line in figure 3 and it clearly indicates that a multi-classifier system with decision fusion performs better than corresponding individual classifiers for recognition complex face images. The maximum recognition rate as 100% has been achieved by this decision fusion technique in two different occasions whereas 81.96% is the average recognition rate for the same method and that is 3% more than that of MLPs, which is a significant improvement so far the performance is concerned.

### CONCLUSION

In this paper, a low level fusion namely data fusion and a high level fusion namely decision fusion have been investigated. Data fusion process has been completed using wavelet decomposition and reconstruction technique. After that images are projected into an eigenspace. Those projected fused eigenfaces are classified separately using a Multilayer Perceptron and Radial Basis Function Neural network. Finally, a classifier combination system for decision fusion has been developed based on Bayesian formulation. Experimental results also show that a combined fusion of two different ANN's gives better result over a single classifier system. This method comprises of wavelet decomposion along with fusions at lower level e.g. data fusion and fusion at higher level e.g. decision fusion is quite effective in combating the detrimental effects of illumination variations. This scheme may be useful for further studies in handling various complicacies exists in face recognition like variations in expressions, aging, pose, beards, moustache, spectacles etc.

## ACKNOWLEDGMENT

Authors are thankful to a major project entitled "Design and Development of Facial Thermogram Technology for Biometric Security System," at Department of Computer Science and Engineering, Jadavpur University, India funded by University Grants Commission(UGC), India for providing necessary infrastructure to conduct experiments relating to this work.